# The State of the Art in Developing Fuzzy Ontologies: A Survey

**Zahra Riahi Samani[1], Mehrnoush Shamsfard[2],**

[1]*Faculty of Computer Science and Engineering, Shahid Beheshti University*

*Email:z_rsamani@sbu.ac.ir*

[1]*Faculty of Computer Science and Engineering, Shahid Beheshti University*

*Email:m-shams@sbu.ac.ir*

**Abstract**

Conceptual formalism supported by typical ontologies may not be sufficient to represent uncertainty information which is caused due to the lack of clear cut boundaries between concepts of a domain. Fuzzy ontologies are proposed to offer a way to deal with this uncertainty. This paper describes the state of the art in developing fuzzy ontologies. The survey is produced by studying about 35 works on developing fuzzy ontologies from a batch of 100 articles in the field of fuzzy ontologies.

# 1. Introduction

Ontology is an explicit, formal specification of a shared conceptualization in a human understandable, machine-readable format. Ontologies are the knowledge backbone for many intelligent and knowledge based systems [1, 2]. However, in some domains, real world knowledge is imprecise or vague. For example in a search engine one may be interested in "an extremely speedy, small size, not expensive car". Classical ontologies based on crisp logic are not capable of handling this kind of knowledge. Fuzzy ontologies were proposed as a combination of fuzzy theory and ontologies to tackle such problems.

On the topic of fuzzy ontology, we studied about 100 research articles which can be categorized into four main categories. The first category includes the research works on applying fuzzy ontologies in a specific domain-application to improve the performance of the application such as group decision making systems [3] or visual video content analysis and indexing [4]. The ontology development parts of the works in this category were done manually or were not of much concentration. The second category includes the researches on developing fuzzy ontologies by an automatic or semi–automatic approach. For example Lee and colleagues [5] propose a method to fuzzify an existing domain ontology and apply it in news summarization. The third category focuses on representation, reasoning and developing inference services in fuzzy ontologies. Most of the researchers in this category offer fuzzy description logic as a theoretical counterpart of fuzzy ontologies [6] and some other researchers propose other approaches [7-9]. The fourth category provides some facilities for fuzzy ontologies. For example Truong and Nguyen [10] propose a framework for fuzzy ontology alignment. Bobillo and Straccia [11] provide the syntax and semantic of a fuzzy description logic with fuzzy aggregation operators.

This paper includes a survey on the state of the art in developing fuzzy ontologies. Thus, our research mainly includes a survey on the second category of researches, although we will introduce some related works in other categories too.

Shamsfard and Abdollahzadeh [12] have introduced a framework to study ontology learning systems. Their framework contains six dimensions; elements learned, starting point, preprocessing, learning method, the result and evaluation method. In this paper we extend their framework and use the new dimension set to study the development (learning) of fuzzy ontologies. In this extension, we've added a new fuzzy related dimension; "supporting fuzziness"; changed the first dimension – learned elements - into a new one: "fuzzified elements" and changed the dimension "developing and test environment" into "application" dimension.

In this paper, after discussing some definitions we describe some fuzzy ontology learning (acquisition and development) research activities and compare them upon our extended framework.

The rest of this paper is organized as follows. In the next section fuzzy logic and fuzzy set theory are briefly introduced. Then we discuss the need to fuzzy ontologies, give some definitions for fuzzy ontologies, and discuss about how to represent them. Then we will have a comparative review on different methods for developing fuzzy ontologies. Finally, we will have a look at the applications of fuzzy ontologies and evaluating them and finally conclusions and open challenges are discussed.

## 2. Fuzzy Logic

Fuzzy set theory, an extension of traditional set theory, is used to represent vague or imprecise information. While in classical set theory elements either belong to a set or not, in fuzzy set theory elements can belong to a set to some degree. Fuzzy set theory uses a membership function to allow belongings of an item in a set to be any real number between 0 and 1. A fuzzy set A with respect to a universe of discourse M is characterized by the membership function μ and a membership value in the unit interval [0, 1], $A = \{x, \mu_A(x)) \mid x \in M\}$ 1 while $\mu_A(x)$ models degree of belongings of x to set A [13].

In fuzzy logic applications the numerical values of variables are usually substituted by non-numeric linguistic variables to facilitate the expression of rules and facts[14]. For example, a linguistic variable such as height for a human may have values such as short, medium height or tall with the following membership functions

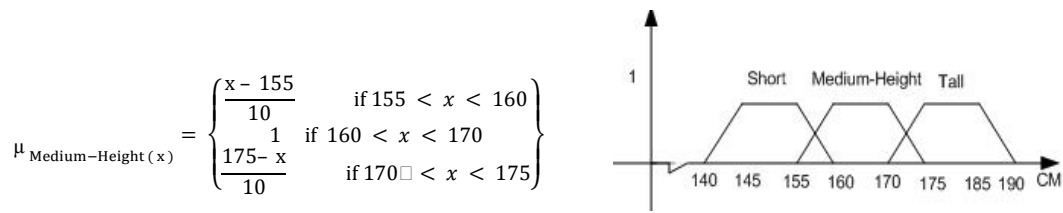

$$\mu_{\text{Medium-Height}(x)} = \begin{cases} \frac{x - 155}{10} & \text{if } 155 < x < 160 \\ 1 & \text{if } 160 < x < 170 \\ \frac{175 - x}{10} & \text{if } 170 < x < 175 \end{cases}$$

**Figure 1. Fuzzy linguistic variable "height" with linguistic values "Short"," Medium height" and Tall for concept human**

However, the great utility of linguistic variables is that they can be modified via linguistic modifiers such as very, slightly and extremely applied to the primary terms. The linguistic modifiers are used to alter the strength of original linguistic values.

Fuzzy logic, based on fuzzy set theory has been applied to several areas such as information retrieval [15], image processing [16] and controller systems [17, 18] which are essential parts to regulate the behavior of sensitive devices

[19, 20]. In this paper, we discuss using fuzzy logic in ontologies to represent knowledge in a domain with vague concepts.

## 3. Why Fuzzy Ontology Is Needed?

Classical Ontologies have some limitations for representing knowledge in real world. Firstly, concepts which are used to describe real world in some domain are inherently vague. For example, a town is a small and less crowded city, but small and crowded are not clearly well defined. It may be difficult to say that a special region fully belongs to the category of towns or cities. It is more desirable to state that it belongs to both of these concepts with different degrees.

Similarly, for having a concept hierarchy it is more rational to have a fuzzy concept hierarchy where a town can be considered as a sub-concept of a city to some degree.

Furthermore, a concept may have attribute values taking vague values, for example a person can have attribute "age" as a linguistic variable taking values "young", "middle aged" or "old" defined by means of fuzzy numbers.

And finally, in some domains we may need fuzzy relations between entities. For example, in the domain of Molecular Biology a relation can be associated with different biological entities with different degrees of strength. In the work of Abulaish and Dey [21] it is observed that the relation "inhibitor of" occurs frequently between protein molecule and protein complexes while the same relation is rare between two protein molecules.

In addition to the inherent uncertainty that may happen in some real-world domains, uncertainty may happen due to the nature of application. For example in a search engine in the case of overloaded concepts[1], the location of concept may be different for users with different preferences [22]. Also, the uncertainty may be because of differences in experts' conceptualization of a domain. For example, New York Times categorizes news into "World, Us, NY/Region, Business, Technology, Science, Health, Sports, Opinion, Arts, Style and Travel", While CNN categorizes news into categories "Home, World, US, Politics, Entertainment, Health, Technology, Travel, Living, Business, Sport and Time". A one-to-one mapping which may be required by an application like news summarization is not possible without inferring a kind of uncertainty [23].

As a result of these imprecision, we need a way to handle this kind of knowledge. Fuzzy ontology would be a good solution to this problem which we are going to define in the next session.

---

[1] Overloaded concept is a concept which occurs in different part of an ontology having relation to different concepts such as java which happens as a kind of coffee, an island in Indonesia south of Borneo or a programming language.

## 4. What Is Fuzzy Ontology?

Different researchers present different definitions for fuzzy ontology. Most of these definitions are application dependent and are on fuzzifying some ontology elements according to their application needs. In other words, the main difference in defining fuzzy ontologies is related to the difference in the ontology element which is going to be fuzzified. According to literature, ontology elements which are involved in the fuzzification process may include concepts, instances, taxonomic and non-taxonomic relations, properties (attributes) and axioms.

There are two ideas for defining fuzzy concepts. Some researchers [24],[25] define fuzzy concept as a concept which has fuzzy attributes, and some others [26] define fuzzy concept as a fuzzy set on the set of instances. In this study we refer to the former as fuzzy attributes and the latter as fuzzy taxonomic relations (the relation between instances and their super class). Thus, the fuzzy elements of an ontology actually include one or more of attributes, taxonomic relations and non-taxonomic relations.

To have a wide-coverage definition of fuzzy ontologies which is acceptable by the community, let's first have an overview on various definitions (from simple to complex) by various researchers in this field. In this overview we divide the definitions into three main categories; (1) including just fuzzy relations or attributes, (2) including both fuzzy attributes and fuzzy relations and (3) including fuzzy attributes, fuzzy relations and fuzzy axioms.

Some researchers define fuzzy ontology as an ontology with fuzzy relations. The relations which are going to be fuzzified may be taxonomic (as in [24]) or non-taxonomic (as in [5]) or both (as in [25]). As an example we can mention Lam [27] who defines fuzzy ontology as $O_F (v, E, l, \mu_f)$ where v is a finite set of vertices (concepts) and $E \in V \times V$ is a set of edges (relations). The edges E are assigned a continuous fuzzy value ($\mu_F: E \to [0\ 1]$) and a label $l : E \to L$. In this definition the fuzziness is applied to a relation as a weight.

Some researchers define fuzzy ontology as an ontology with fuzzy attributes. Samani and Shamsfard [28],[29] define fuzzy ontology as a 4 tuple $O_F= (C, P^C, R, A)$ where,

- C is a set of concepts.
- $P^C$ is a set of entity properties that can be represented as a 8 tuple $p^c (c, p, v_p, g_p, n_p, q_p, h_p, f)$ where
  - c is an ontology concept.
  - p is the property name.
  - $v_p$ is a set of values of the property.
  - $g_p$ is a set of membership functions assigned to the members of $v_p$.

- $n_p$ is a set of membership degrees assigned to $v_p$.
- $q_p$ models linguistic modifier (which is optional).
- $h_p$ is a set of membership functions assigned to each modifier.
- f is a restriction facet such as type or cardinality. The type may be {Integer, float, etc}. Cardinality defines the upper and lower limits on the number of values of the property.
- R is a set of relations between concepts.
- A is a set of axioms.

They state that according to the reasoning requirement different parts of ontologies may be fuzzy. As in the application of their model they need fuzzy attributes they give this definition.

Some other researchers extend the fuzziness to cover more elements and define fuzzy ontology as an ontology with fuzzy attributes, and fuzzy relations. In this category we can mention Lau [24] who defines fuzzy ontology as a 6-tuple $O_F$ = { X, A, C, $R_{XC}$, $R_{AC}$, $R_{CC}$} where

- X is a set of objects.
- A is the set of attributes describing the objects.
- C is a set of concepts (classes).
- The fuzzy relation $R_{XC}$: X × C →[0 1] assigns a membership to the pair ($x_i$, $c_i$) for all xi ∈ X, ci ∈ C.
- The fuzzy relation $R_{AC}$: A ×C →[0 1] defines the mapping from the set of attributes A to the set of concepts C.
- The fuzzy relation $R_{CC}$: C ×C →[0 1] defines the strength of the sub-class/ super-class relationships among the set of concepts C.

In this definition, $R_{AC}$ is denoting fuzzy attributes (fuzziness in belonging an attribute to a class) and $R_{XC}$ and $R_{CC}$ are showing fuzzy taxonomic relations (instance-of and sub-class-of respectively). This research does not talk about (fuzzy or crisp) axioms.

Abulaish and Dey [30, 31] are other members of this category who give a more descriptive definition. They define fuzzy ontology as an ontology with fuzzy properties and relations and let these fuzzy elements be described by fuzzy numbers or linguistic quantifiers. In their definition a fuzzy ontology is shown by $O_F$ (C, $P_F$, $R_F$, M) Where

- C is a set of concepts.

- $P_F$ is a set of fuzzy concept properties. A property $P_F$ is defined as a quadruple of the form $p_f(c, v_f, q_f, f)$ where
    - $c \in C$ is an ontology concept
    - $v_f$ represents fuzzy attribute values which is either fuzzy number or fuzzy quantifier.
    - $q_f$ models linguistic qualifiers which are modifiers.
    - $f$ is the restriction facets on $v_f$.
- $R_F$ is a set of inter-concept relations. Like fuzzy concept properties, $R_F$ is defined as a quadruple of the form $r_f(c_1, c_2, t, q_f)$ where
    - $c_1, c_2 \in C$ is an ontology concept.
    - $t$ represents relation type.
    - $q_f$ models relation strengths and are linguistic variables which can represent the strength of association between concept-pairs $<c_1, c_2>$.
- M defines universe of discourse which is the range an attribute can take value.

Dey and Abulaish [23] define fuzzy ontology as $O_F(C, R_F)$ Where

- C denotes the set of domain concepts.
- $R_F$ denotes a set of inter-concept relations each defined for each pair of concepts which is defined by a quadruple $R_F = \{r_f | <r_f, (c,d), v_f, q_f>\}$ where
    - r is a relation name.
    - $(c, d) \in C \times C$ represents ordered pair of ontology concepts.
    - $v_f$ denotes fuzzy values that can be associated to the relation.
    - $q_f$ models fuzzy qualifier.

In this research they define two kinds of relations including inter-concept relations and concept descriptors which are like properties, In the formal definition, only fuzzy relations are fuzzy, but conceptually they make properties fuzzy too.

There are also some work which define fuzzy ontology with special kinds of relations, attributes or properties. For example Lee and colleagues [5] define fuzzy ontology as an extended domain ontology with fuzzy concepts and fuzzy relationships. They define two kinds of fuzzy relationships, including Location Narrower Relationship (LNR) and Location Broader Relationship (LBR).

They try to make the news ontology fuzzy. Concepts are refined by embedding a set of membership degrees associated with a set of news events. Although the authors claim to have fuzzy concepts and fuzzy relationships, their paper just talks about fuzzifying part-of (meronymy) relations in a hierarchy.

As another example Calegari and Ciucci [26] define a fuzzy ontology as $O_F = \{C, I, R, F, A\}$ where

- I is the set of individuals.
- C is the set of concepts. Each concept $c \in C$ is a fuzzy set on the domain of instances, $C: I \rightarrow [0\ 1]$.
- R is a set of relations. Each $r \in R$ is an n-ary fuzzy relation on the domain of entities (The set of concepts and individuals), $R: E^n \rightarrow [0\ 1]$.
- F is a set of the fuzzy relations on the set of entities E and a specific domain contained in D = {integer, string...}. In detail, they are n-ary functions such that each element $f \in F$ is a relation $F: E^{(n-1)} \times P \rightarrow [0\ 1]$ where $P \in D$.
- A is the set of axioms expressed in a proper logical language.

However, they previously have defined fuzzy ontology with Fuzzy instances and properties [32] in a simpler way. In their older definition, a fuzzy ontology is an ontology extended with fuzzy values which are assigned through the two functions

$$g: (Concepts \cup Instances) \times (Properties \cup Prop_{val}) \rightarrow [0\ 1]$$

$$h: (Concepts \cup Instances) \rightarrow [0\ 1]$$

Ghorbel and colleagues [33] is another definition of this category which defines a fuzzy ontology as a 7-tuple $O_F = (C, P, C_F, P_F, R, R_F, As, As_F, A)$ where:

- C is a set of crisp concepts defined for the domain.
- P is a set of crisp concept properties.
- $C_F$ is a set of fuzzy concepts. A fuzzy concept is a concept which possesses, at least, one fuzzy property
- $P_F$ is a set of fuzzy concept properties. A fuzzy property is a property which is represented in the form of fuzzy linguistic variable.
- R is a set of crisp binary semantic relations defined between concepts in C or fuzzy concepts in $C_F$.
- $R_F$ is a set of fuzzy binary semantic relations defined between crisp concepts in C or fuzzy concepts in CF. A fuzzy binary semantic relation is a relation which be represented in the form of a fuzzy linguistic variable.

- As is a set of crisp binary associations defined between concepts in C or fuzzy concepts in $C_F$.
- $As_F$ is a set of fuzzy binary associations defined between crisp concepts in C or fuzzy concepts in $C_F$. A fuzzy binary association is a relation which be represented in the form of a fuzzy linguistic variable.
- A is a set of axioms. An axiom is a real fact or reasoning rule.

The last category of researches involves axioms in fuzzy ontology definition as well as properties and relations. For example Zhai and colleagues [34] present the most covering general definition. They define fuzzy ontology as $O_F$ (I, C, $P^C$, R, $P^R$, $A_F$) where

- I is the set of individuals, also called instances of the concepts.
- C is a set of concepts.
- $P^C$ is a set of concepts properties. A property $p \epsilon P^C$ is defined as a 5-tuple of the form $p^c$ (c, $v_f$, $q_f$, f, U) where
    - $c \epsilon C$ is an ontology concept,
    - $v_f$ represents property values,
    - $q_f$ models linguistic qualifiers, which can control or alter the strength of a property value $v_f$,
    - f is the restriction facets on $v_f$, and
    - U is the universe of discourse.
- The property $p^c \epsilon P^C$ has also the non-fuzzy form $P^C$ (c, v, f).
- R is a set of inter-concept relations between concepts. The relation type is not only the ordinary binary relation of $r \epsilon C \times C$, but also is the fuzzy relation from C to C.
- $P^R$ is a set of relations properties. Like concept properties, $p^r \epsilon P^R$ is defined as a 4-tuple of the form $p^r$( $c_1, c_2, r, s_f$ ) where
    - c1, c2 $\epsilon C$, are ontology concepts,
    - r represents relation, and
    - $s_f$ :[0,1] models relation strengths and has meaning of fuzzy set on $C \times C$, which can represent the strength of association between concept-pairs ( c1 , c2 ) .
- $A^F$ is a set of fuzzy rules.

While in Abulaish and Dey's [30, 31] definition, relations and properties may be either fuzzy numbers or fuzzy linguistic variables, Ghorbel and colleagues [33] define both of them as a linguistic variable and in Zhai et al's [34] definition, relations are fuzzy numbers and properties may be either fuzzy numbers or fuzzy linguistic variables. Zhai et al [34] also includes fuzzy rules which are not present in the definition of Abulaish and Dey [30, 31] and Ghorbel and colleagues [33].

Ortega [35] defines fuzzy ontology as an ontology which uses fuzzy logic to provide a natural representation of imprecise and vague knowledge and ease reasoning over it. He disagreed to define a fuzzy ontology by enumerating its fuzzy elements as it threats the scalability and reusability of definitions. He states that new languages will offer new possibilities to be extended with fuzzy elements but current definitions do not cover them. For example, none of the above definitions mentioned fuzzy taxonomy of relations.

However, if we want to give a formal definition for fuzzy ontology there is no other way but enumerating fuzzy elements. As it can be seen there are many common components in different definitions given for fuzzy ontologies. In this paper, we complete the previous definitions to cover various dimensions of fuzziness in fuzzy ontologies. To do so, we modify and extend the definition of Samani and Shamsfard [28] to fuzzy relations and define fuzzy ontology as a 5 tuple $O_F = (C, P^C, R, P^R, A_F)$ where,

- C is a set of entities.
- $P^C$ is a set of entity properties that can be represented as a 8 tuple $p^e(e, p, v_p, g_p, q_p, h_p, f)$ where,
    - E is an ontology concept.
    - p is the property name.
    - $v_p$ is a set of values of the property.
    - $g_p$ is a set of membership functions assigned to the members of $v_p$.
    - $q_p$ models linguistic modifier (which is optional).
    - $h_p$ is a set of membership functions assigned to each modifier which shows how the fuzzy value for a combination of this modifier with the base linguistic value should be computed.
    - f is a restriction facet such as type or cardinality. The type may be {Integer, float, etc}. Cardinality defines the upper and lower limits on the number of values of the property. Other facets may be defined by the ontology engineers.
- R is a set of relations.
- $P^R$ is a set of relation properties. Such as $P^C$, $P^R$ is a 8 tuple $p^r(c_1, c_2, r, v_r, g, q_r, h_r, f)$ in which
    - $c_1, c_2 \in C$ are concepts of ontology.

- r represents a relation between $c_1$ and $c_2$.
- $v_r$ is a set of linguistic values for the relation value.
- $g_r$ is a set of membership functions assigned to each linguistic value.
- $q_r$ models a set of linguistic modifiers of linguistic values.
- $h_r$ is a set of membership functions assigned to each modifier.
- F is a restriction facet like domain and range.

- $A_F$ is a set of fuzzy axiom in the form (P, $n_t$). Where P is a proposition like (IF A THEN B) and $n_t$ is its truth degree.

Classical ontologies are special kind of fuzzy ontologies. So, we want this definition to be backward compatible with classical ontologies. In classical ontologies a concept property is defined by a triple (c, p, f) and a relation is defined by triple ($c_1$, $c_2$, f) with other variables as an empty set.

For example, in an ontology concept: Human has the property: height ($P^C$) which can have linguistic values short, medium-height and tall ($v_p$) with their assigned membership function ($g_p$) (as in Figure 1).

A set of modifier such as very and slightly ($q_p$) may be assigned to the base linguistic values with their assigned membership functions ($h_p$). For example, one possible $h_p$ for modifier very could be defined in this way:

$$\mu_{Very-X} = (\mu_X)^2 \text{ where } X \text{ is the linguistic value like Small}$$

Suppose in an ontology the concept human be a child of a parent concept "living things". So, the $n_p$ {0, 0.4, 0.6} shows that the concept human has the membership 0.4 to the medium-height living thing and membership degree 0.6 to the tall living thing. The height attribute is represented formally in the following way.

{E: "Human", P: "Weightt", $v_p$:{ light, Medium-weight, Heavy}, $g_p$:{ 140-145-155-160, 155-160-170-175,170-175-185-190}, $n_p$:{ 0,.04,0.6}, $q_p$:{ very}, $h_p$:{ Pow-By-2}, f:{ Range: (135-185)}}}

The same example is held for the fuzzy relation. For example, in an ontology one may need to assign strength associated to its relations. This strength can be a number or a linguistic value such as strong, weak and medium and optionally a set of modifiers can be assigned to it with their own linguistic value. Also, it is possible to talk about other properties of relation (a reification process may be needed). For example, we may need to enrich the ontology with the fuzzy attribute price of the relation "hiring". Linguistic values like cheap, medium, expensive can be defined. In the next Section we will discuss about representing fuzzy ontologies.

## 5. How Fuzzy Ontology Is Represented?

As the theoretical counterpart of fuzzy ontology, the fuzzy description logics (FDLs) have also attracted much attention from researchers.

Concepts and relations in fuzzy description logic are fuzzy sets. For example, the concept "YoungMan" is a fuzzy concept and is defined as:

$$YoungMan = Man \cap \exists Age.Young$$

"age" expression is a fuzzy concept (linguistic variable) and it gets fuzzy linguistic value:" young". Fuzzy linguistic value "young" may be defined with a trapezoidal function such as

$$\mu_{Young}(c.Age) = \begin{cases} 1 & if B \leq C.Age \leq C \\ \frac{D-C.Age}{D-C} & if C < C.Age < D \\ \frac{C.Age-A}{B-A} & if A < C.Age < B \\ 0 & Otherwise \end{cases}$$

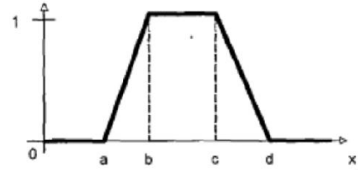

**Figure 2-Fuzzy trapezoidal function for concept "YoungMan".**

Figure 2 shows the diagram of the trapezoidal function. Fuzzy Description Logic extends the syntax and semantic of concepts and roles to the fuzzy ones by adding fuzzy membership weights to define a new syntax and semantic on the crisp DLs. A membership function (MF) is a curve which illustrates the mapping between the input space to a membership value {between 0 and 1}[36]. There are different forms of membership function available in literature like triangular, trapezoidal, Gaussian, singleton, and piecewise linear[37].

A lot of researchers have studies fuzzifying the existing description logics. For examples, Straccia [38] presented fuzzy ALC and a fuzzy version of SHOIN (D) [39]. Stoilos and colleagues [40] introduced a fuzzy extensions of SHIN and a fuzzy SHOIN [41]. Bobillo and Straccia [42] provide the syntax and semantics of a fuzzy Description Logic with fuzzy integrals. For a survey on Fuzzy DLs readers can refer to [43].

However, not all researchers talk about fuzzy DLs for representing fuzzy ontologies. Bobillio and Straccia [44] propose to have an ontology to represent concepts of a fuzzy ontology and use this ontology to represent fuzzy ontologies within current Semantic Web languages. In another research, Bobillio and Straccia [45] use OWL2 annotations for representing fuzziness in ontologies. Lv and colleagues [7, 8] propose fuzzy relational database as an efficient solution to store fuzzy ontologies. Yeung and Leung [9] propose a formal model for handling fuzzy membership and typicality of Instances in ontologies. Then a logic for the ontology model based on fuzzy propositional modal logic is presented. Cai and Leung [46] propose a formal model for fuzzy ontology with the ability of property importance and property hierarchy.

Technologies of ontologies have a special role in semantic web. For representing fuzzy knowledge in semantic web, fuzzy extension of semantic web languages such as RDF [47], OWL [48], RuleML[49] ,etc are also presented.

Besides representing fuzzy concepts and roles, development of fuzzy description logic has provided the capability of reasoning with fuzzy data. Stoilos and colleagues [40] introduced a tableaux algorithm for the fuzzy extensions of SHIN and also the fuzzy extensions of f-SHOIN [41]. Straccia [38] presented an algorithm for reasoning on fuzzy ALC. The algorithms are based on expansion rules defined in fuzzy DL. These Expansion rules transform a set of fuzzy constraints to a set of simpler ones until finding a clash or making a clash free set of constraints.

Several reasoners for fuzzy description logic is also provided such as FIRE [50] which implements a tableaux algorithm for fuzzy SHIN. Delorean [51] which reduces reasoning in fuzzy SHOIN to reasoning in crisp SHOIN, And Fuzzy DL [52] which implements a combination of a tableaux algorithm and a MILP optimization for the DL SHIF. Others include GERDS [53],GURDL [54],YADLR [55],etc.

## 6. How Fuzzy Ontology Is Developed?

In this section, we divide developing of fuzzy ontologies into three parts including methodologies, models and frameworks and developing methods. Methodologies are used to develop the fuzzy ontology manually. Models and frameworks need assignment of fuzzy weights manually by domain experts. They offer the framework to reason new fuzzy memberships from the initial existing fuzzy memberships or provide some application based on their models. Developing methods use automation to provide fuzzy memberships. However, in some parts human intervention may be needed. We name methods without special name with the name of their authors[2]. The next section compares these methods.

### 6.1. Fuzzy Ontology Development Methodologies

As we stated, there are some researches which provides methodologies. Some researchers use their own methodologies for developing fuzzy ontology [33, 56-60]. From them the following is selected which was the most comprehensive one with least dependency to the domain.

---

[2] In this Paper, from papers written by the same authors with similar content, we chose one of them which were most recent or most cited.

### 6.1.1. IKARUS-Onto

IKARUS-Onto (Imprecise Knowledge Acquisition Representation and Use) [60] propose a methodology for developing fuzzy ontologies from existing crisp ones. The focus of the methodology is not so much on the structure of the fuzzy ontology but rather on the process of its development and the ultimate content it has. It has some tips and guide lines for ontology engineers to correctly identify the vague knowledge of the domain, determine fuzzy ontology elements and decide about the degree of membership values.

In this methodology, vagueness is defined as predicate that contains borderline cases. Borderline cases are cases where it is unclear if the predicate applies. For example, a person is borderline tall: not clearly tall and not clearly not tall. According to their definition, two basic kinds of vagueness are defined: degree vagueness and combinatory vagueness. Degree-vagueness occurs when the existence of borderline cases stems from the lack of precise boundaries along some dimensions. For example, tall has not a sharp boundary along dimension height. Combinatory vagueness occurs when there are a variety of conditions but it is not possible to say which are sufficient for application and which are not. An example of this type is Religion. There are certain features that all religions share (e.g. beliefs in super natural beings), but it is not clear which of these features are enough to classify something as a religion. The methodology has the following steps.

**Step 1: establishment of the need for fuzziness:** Establishing the need for fuzziness means determining whether and to what extent is vagueness present in the domain at hand. A concept or relation is vague, if in the given domain, or application, it admits borderline cases, means if there are individuals for which it is indeterminate whether they instantiate the concept or if there are pairs of individuals for which it is indeterminate whether they stand in the relation. The same applies for attributes, pairs of individuals and literal values.

**Step 2: definition of fuzzy ontology elements:** Through this step, the nature of the fuzzy element's vagueness and the expected interpretation of its fuzzy degrees are made explicit. The goal of this description is to ensure that the defined fuzzy elements have a clear and specific vague meaning which makes them shareable and reusable. For this purpose, it has steps for defining fuzzy attribute values and relations, fuzzy data types and fuzzy concepts. The following tips are given for specifying fuzzy relations and attributes.

1. Determining for each relation/attribute, the type of its vagueness (combinatory or degree-vagueness). If the element has degree-vagueness, then the dimensions along which it is vague need to be identified.

2. Defining for each element the exact meaning of its vagueness. If the element has degree-vagueness along multiple dimensions, then the distinction between the dimensions might or might not be important. In case it is then, it would be necessary to define a distinct fuzzy element for each dimension.

3. Defining of the expected interpretation of each element's fuzzy degrees. If fuzziness is due to degree-vagueness, then the fuzzy degree of a related pair of instances means the extent to which the pair's value for the given dimension places it within the elements' application boundaries. If fuzziness is due to combinatory vagueness, then the fuzzy degree approximates the extent to which the pair's set of satisfied application conditions of the relation/attribute is deemed sufficient for the relation/attribute to apply.

4. The assignment of specific fuzzy degrees to pairs of instances (or instances and literal values) that instantiate each element.

The process followed for the definition of fuzzy concept is similar to fuzzy relations and attributes with an important difference. In many cases, vague concepts "owe" their vagueness to some vague relation, attribute or term which has been already defined. If that is the case, then the definition of the concept's vagueness can be directly derived from the one of its causal element.

**Step 3: formalization of fuzzy ontology elements:** this step some tips to decide about the way of representing fuzzy ontology. For this decision, the range of fuzzy ontology elements, and the range of fuzzy reasoning capabilities it support should be considered as representing languages has different expressive power and different reasoning capabilities.

**Step 4: validation of fuzzy ontology:** This method defines the following criteria for the evaluation of fuzzy ontology

1. Correctness: A fuzzy ontology is correct when all its fuzzy elements convey a meaning which is indeed vague in the given domain or application.

2. Accuracy: A fuzzy ontology is accurate when the degrees of its fuzzy elements approximate the latter's vagueness in an intuitively accurate way for the given domain or application.

3. Completeness: A fuzzy ontology is complete when all the vagueness of the domain has been represented within the ontology.

4. Consistency: A fuzzy ontology is consistent when it does not contain controversial information about the domain's vagueness as this is expressed by fuzzy degrees.

This methodology has been applied in the developing a fuzzy enterprise ontology for a consulting firm.

## 6.2. Fuzzy Ontology Models and Frameworks

### 6.2.1. Three-Layer Fuzzy Ontology Model

Zhai and colleagues [34] propose a three-layer architecture (model) for fuzzy ontologies. The layers include: concepts, properties of concepts and values of properties. Property values are ordinary values or linguistic values of fuzzy concepts which are defined by a fuzzy linguistic variable ontology. The fuzzy linguistic variable ontology contains all the property values with their assigned membership functions and the relation between them. It is defined by domain experts. The model is the extension of RDF data model "object-property-value".

Figure 3 shows an example of this model in an e-commerce application. This ontology has concept customer which has three attributes Age, Type and Income. Each attribute has its own linguistic values which are defined by domain experts.  And relations like order relation such as youth≤middle-aged≤young, equivalence relations like "gold customer"= "big customer", or inclusion relation like "switched customer" ⊆ "lost customer can be defined. Others include reversion relation and complement relation.

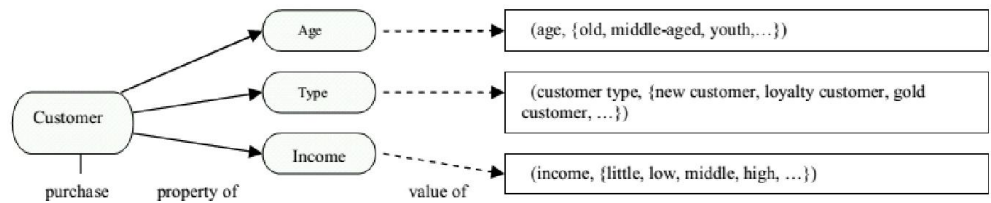

**Figure 3-An example of 3-layer architecture for of e-commerce application [34].**

A semantic query expansion process is constructed by the defined relations between fuzzy concepts of linguistic variable ontology. For instance, the "product" information through the property "price" can be retrieved using the search statement such as:

SELECT Product (name, brand, price, …) FROM Data source WHERE Product.price ≤ expensive".

They state that the standard ontology and other fuzzy ontologies are not able to handle the search condition at semantic level, which includes fuzzy concept and semantic relation between them.

### 6.2.2. Samani and Shamsfard

Samaniand and Shamsfard [28],[29] propose a fuzzy ontology model for fuzzifying attribute values. They propose a complex data property structure (shown in figure 4) for keeping fuzzy attributes. The complex data property includes a linguistic variable and two parts of information about it, the crisp part and the fuzzy part. The crisp part (CrispInfo) contains crispValue which shows the ranges of the crisp value for *concepts* and the real crisp value (if any) for *instances*. It also contains the unit of measurement with which the value is measured. The fuzzy part includes linguistic values and the modifiers. Linguistic value and modifiers get value in the concept (they are concept attributes) while linguistic values membership degree get value in the instances (is instance attribute).

The other parts of FuzzyInfo are the modifiers. Each modifier contains 3 fields including modifier-name, modifier type (expansive or restrictive), and modifier shifting number which shows how the membership function of the modified linguistic value could be computed. As an example the complex data property for the attribute "Size' of the concept "LapTop" is shown in (Figure 4 right) They also provide a reasoning mechanism based on the proposed fuzzy ontology model for qualitative reasoning. They state that if the attribute gotten fuzzy is spatial attribute like "width", "height", "length", etc, the method is capable of doing qualitative spatial reasoning. They propose an extended version of OWL called E-OWL for representing their model.

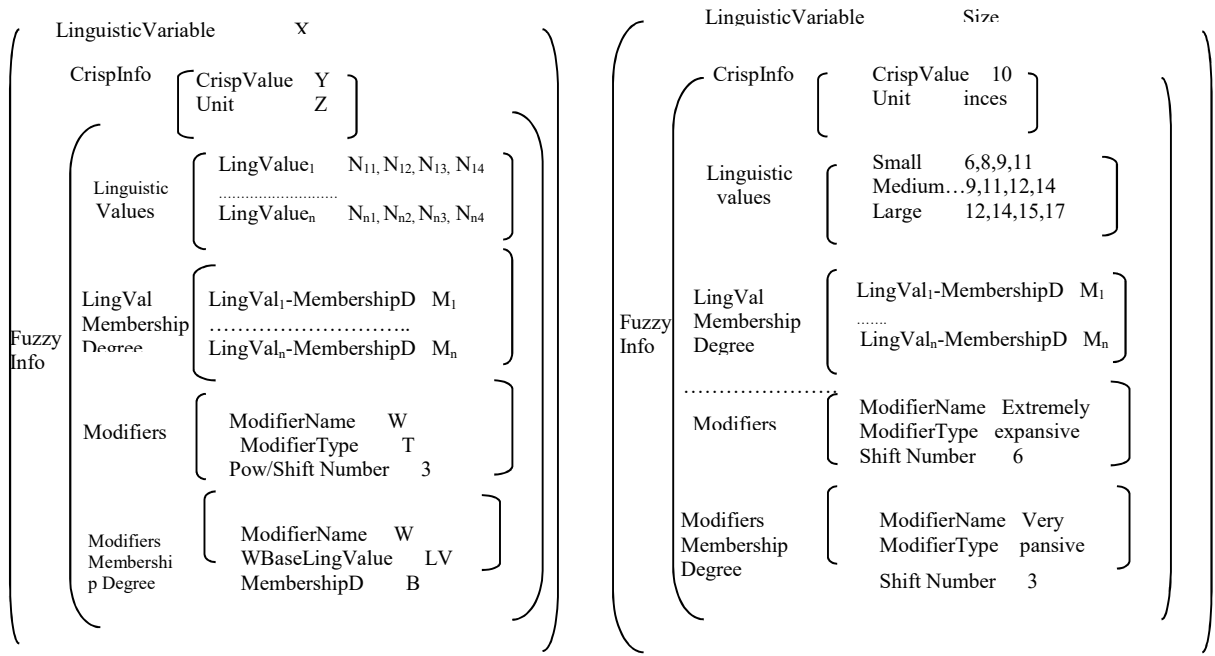

**Figure 4 Structure of Complex Data Property (Left)-A sample for concept laptop(Right)[28]**

### 6.2.3. FOM

Lam [27] proposes Fuzzy Ontology Map (FOM) as an extension of a current crisp ontology. FOM is based on fuzzy theory and graph theory, it is a connection matrix which collects the membership value between classes in the ontology. $i^{th}$ row in FOM represents fuzzy weights $e_{ik}$ for class $A_i$ and jth column represents fuzzy values $e_{kj}$ to class $A_j$. It starts with a base matrix created by domain experts. Starting from this base, inferring algorithms for creating all virtual edges for all reachable and unreachable classes is represented. Reachable and unreachable classes are defined in this way.

- *Reachable Classes:* A, B, C are the classes in FOM. If there is an edge $e_{BA}$ connected from class B to class A with membership value $\mu_{BA}$ and an edge $e_{CB}$ connected from class C to class B with membership value $\mu_{CB}$. As a result, class A is reachable from class C by class B. So if there is no edge from class C to class A, a virtual edge $e_{CA}$ is derived with membership value $\mu_{CA}$.

$$\mu_{CA} = \mu_{CB} \times \mu_{BA}$$

- *Unreachable Classes:* A, B, C are the classes in FOM. If there is an edge $e_{BA}$ connected from class B to class A with the membership value $\mu_{BA}$ and an edge $e_{CA}$ connected from class C to class A with the membership value $\mu_{CA}$ but there is no edge between class B and class C (either from class B to class C or from class C to class B), then two conditions are discussed
  - If the fuzzy relations are symmetric, virtual edges $e_{BC}$ and $e_{CB}$ is derived with the membership value $\mu$.

$$\mu = sim(\mu_{BA}, \mu_{CA}) = \frac{\min(\mu_{BA}, \mu_{CA})}{\max(\mu_{BA}, \mu_{CA})}$$

  - If the fuzzy relations are a symmetric $e_{CB}$ is derived with the membership value $\mu_{BC}$

$$\mu_{BC} = asymm(\mu_{BA}, \mu_{CA})) = \frac{|\mu_{BA} - \mu_{CA}|}{\mu_{CA}} \qquad \mu_{CB} = asymm(\mu_{CA}, \mu_{BA})) = \frac{|\mu_{CA} - \mu_{BA}|}{\mu_{BA}}$$

If A and B are classes in FOM and there is an edge starting from class A and ending at class B with the membership value $\mu_{AB}$ and there is no edge in the opposite direction, if the fuzzy set is symmetric, a virtual edge from class B to class A can be created with the membership value $\mu_{symm}$,

$$\mu_{symm} = \mu_{AB}$$

The method proposes to store fuzzy ontology using two files: an RDF/OWL document for the domain concept hierarchy and an XML document for the fuzzy information. The performance of creating FOM is evaluated and some recommendations for having a better performance are offered.

### 6.2.4. Gu And Colleagues

Gu and colleagues [61] present a framework for developing a fuzzy ontology which is general (not domain specific) and supports reasoning. Their method introduces three relations which are found in all domains. The relations are called fuzzy instance relation, fuzzy concept relation and fuzzy concept base relation. Fuzzy instance relation is ordinary role in fuzzy description logic. Based on this relation, others are defined.

Fuzzy concept relation: In some domains, there may exist some special fuzzy instance relations such that, all the membership degrees between the instances of two concepts are the same. So the relation is moved to the concepts and is called fuzzy concept relation (Figure 5-left).

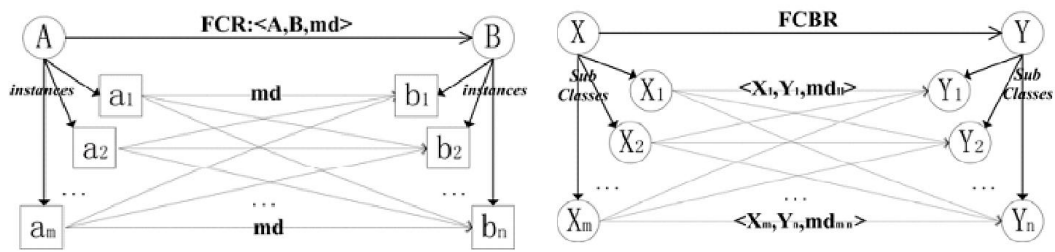

**Figure 5-Fuzzy Concept Relation (left) Fuzzy Concept Base Relation(right) [61]**

Fuzzy concept base relation (FCBR for short): It is defined as a set of fuzzy concept relations such as the concept set $\{X_1, X_2…, X_m\}$ of X and the set $\{Y_1, Y_2…,Y_n\}$ of Y (the relation between $X_i$s and X and $Y_i$s and Y are "kind of" ) such that, the membership of all of the $X_i$s in X and $Y_i$s in Y are the same. So the relation is moved to the parent concept (Figure 5-right).

A reference process for guiding the construction of fuzzy concept base relations is then introduced. The process is semi-automatic and needs human intervention. Having the candidate FCBR relations (X, Y), The process recursively uses all the kind of relations $X_i$s of X and $Y_i$s of Y and builds a matrix for them. Representing this matrix to the domain experts and asking for the suitable fuzzy weight assignment is the next step.

An extended version of fuzzy description logic called ef-shin is introduced. Mapping of these three relations to ef-shin is discussed. Because of the mapping provided, the method claims that it supports fuzzy reasoning.

## 6.3. Fuzzy Ontology Developing Methods

As it can be seen in the previous section, fuzzy ontologies may have some fuzzy elements including fuzzy concepts, attributes, relations and axioms. In this section we described some popular learning methods used to extract the fuzzy elements.

### 6.3.1. Extracting Fuzzy Attribute Value

#### 6.3.1.1. Abulaish and Dey(A)

Abulaish and Dey [23] propose a fuzzy ontology generation framework in which a concept descriptor is defined as a fuzzy relation which encodes the degree of a property value, using a fuzzy membership function. This framework proposes to store concept descriptions in a <property, value, qualifier, constraints> form as an extension of the traditional <property, value, constraints> framework. Qualifiers are modifiers or hedges and are used to dynamically create new fuzzy sets and change the meaning of linguistic variables. They are extracted through text-mining or are defined by domain expert. A fuzzy mechanism for integrating new qualifiers to the set of original ones is also presented.

In Dey and Abuliash [31] this framework is used to enhance an existing crisp ontology with fuzzy property descriptors gotten through rule based in combination with NLP text mining.

Gathering the information to enhance an existing ontology needs to locate concepts, properties and relation from free-form text. For this purpose, some rules are defined. In the rules, adjectives model properties, adverbs model qualifiers and verbs are relations between concepts. However, there could be numerous such elements in any given text obeying these rules, not all of them necessarily are relevant, so lexical patterns are identified for recognizing ontological concepts from text. They employ SPAM algorithm to mine such structures from annotated text. Patterns like "determiner, adjective, noun" or "noun, verb, preposition, noun" are defined. So first they find components which match the patterns and second they try to find appropriate match in the ontology. If a match is not found, the pattern is stored for verification. Otherwise, the pattern is accepted as an information component.

The proposed fuzzy ontology structure is applied in a text information retrieval application. For matching a pair of <value, qualifier> tuples, the overall effect is also influenced by the distance between the qualifiers as it is influenced by the distance between value pairs.

### 6.3.2. Extracting Fuzzy Relations

In this section we divide methods into two categories. Methods that make fuzzy taxonomic relations and methods that make fuzzy non-taxonomic relations.

#### 6.3.2.1. Fuzzy Taxonomic Relation
The following two methods are discussed.

### 6.3.2.1.1. Lau and PASS

Widyantoro and Yen [62] describe the PASS (Personalized Abstract Search Service) system. The system uses a fuzzy ontology of term associations to support information retrieval. The fuzzy ontology is automatically built using information obtained from the system's document collection. The system first extracts a set of two or three consecutive words with patterns like (noun/adj) (noun/noun noun) or (adjadj noun). It uses WordNet to tag each word. Then the system eliminated two or three word phrases which include at least one word not contained in a predefined control list. Making fuzzy taxonomic relations between terms is done next. Two taxonomic relations narrower than ($NT(t_i,t_j)$) and broader than ($BT(t_i, t_j)$) are introduced and the fuzzy conjunction operator of term frequencies is applied to compute the membership values of the term relations in the following manner.

Let $C = ( a_1 , a_2 ,a_3 , …, a_n )$ be a collection of articles, where each article $A= (t_1,t_2, t_3,..)$ is represented by a set of terms $t_j$ and let $occur(t_j, a)$ denote the occurrence of term $t_j$ in article a. The membership degree of $\mu_{occur}(t_j,a)$ is defined by $f(|tj|)$

$$\mu_{Occure}(t_j,a)=\mu(t_j,a)=f([t_j])$$

where f is defined with term's frequency of occurrence. The membership of $\mu_{NT}(t_i, t_j)$ is defined in this way

$$\mu_{NT}(t_i, t_j) = \frac{\sum_{a \in C} \mu_{occur(t_i,a)} \cup \mu_{occur(j,a)}}{\sum_{a \in C} \mu_{occur(t_i,a) \cup}}$$

And the following equation is used.

$$\mu_{BT}(t_i,t_j) = \mu_{NT}(t_j,t_i)$$

Two additional steps for eliminating redundant relations are done. In the first step, the system does α cut on relations. Then, from the two $NT(t_i, t_j)$, $BT(t_i, t_j)$ relations, the relation with smaller value is eliminated. In the second step, looking for indirect path $NT(t_i,t_{m1})$, $NT(t_{m1},t_{m2})$, …., $NT(t_{mn},t_j)$ is done. If the path is find, on the condition that the value of $NT(t_i,t_j)$ is less than the minimum value in the finding path, the relation $NT(t_i,t_j)$ is eliminated without losing any information..

The work of Lau [24] is another research which makes a taxonomy of concepts. The method is very much similar to the method adapted in PASS. The difference is that this method uses a context vector for representing each concept.

The method does a pre-processing step which includes Stop words removal, part of speech tagging, stemming. Then, a windowing process with term size 5 to 10 is conducted over the corpus. Some patterns like "Noun Noun" and "Adjective Noun" are defined for filtering noisy patterns. After parsing the whole corpus, a statistical token analysis step use information theoretic measure to compute statistics of the linguistic patterns.

Several measures including Mutual Information (MI), Jaccard (JA), conditional probability (CP), Kullback-Leibler divergence (KL), and Expected Cross Entropy (ECH) are introduced for computing context vector of concepts. These statistics are used to define potential concepts. If the association weight between a concept and a term is below a predefined threshold value, the term will be discarded from the context vector of the concept.

Finally, fuzzy memberships for the taxonomy of domain concepts are assigned according to the fuzzy conjunction operator (like defined for PASS) over the terms of the concepts' context vector. Let Spec($c_x$,$c_y$) denote that concept $c_x$ is a subclass of another concept $c_y$. The degree of such a specialization relation can be estimated from

$$\mu_{R_{CC}}(C_x, C_y) \approx Spec(C_x, C_y) = \frac{\sum_{t \in C_z \cap C_y} \mu_{c_z}(t) \otimes \mu_{cy_z}(t)}{\sum_{t \in C_z} \mu_{cy_z}(t_x)}$$

Similar approached is applied in [63-65].

### 6.3.2.1.2. Chien and Hsu

Chien and colleagues [66] propose an agglomerative clustering scheme based on fuzzy theory to generate hierarchical fuzzy concepts from a large database automatically.

The proposed method generates hierarchy of concepts with the number of layers equal to the number of attributes in the data base. The method starts with database entities as the last layer of the hierarchy. Then a clustering algorithm is applied to the primary objects to find meaningful fuzzy concept hierarchies effectively.

The first part of the method is transforming quantitative data in database entities into linguistic terms using fuzzy membership functions defined for their attributes. Figure 6 shows the membership temperature and humidity. Table 1 shows a snap-shot of an example database with 5 attributes outlook, temperature, humidity, windy and play.

As the number of attributes is 5, the hierarchy has 5 layers. In the lowest layer concept $C^5_i$ represents a direct entity $x_i$ in database. For example

$C_1^5$= ((0/O+0/R+1/S), (1/H+0/M+0/L), (0.65/H+ 0.35/L), (0/T+1/F), (0/Y+1/N))

which shows that object $C_1^5$ has value 1 for linguistic value high, and value 0 for linguistic values medium and low for attribute temperature. Other values have the same interpretation.

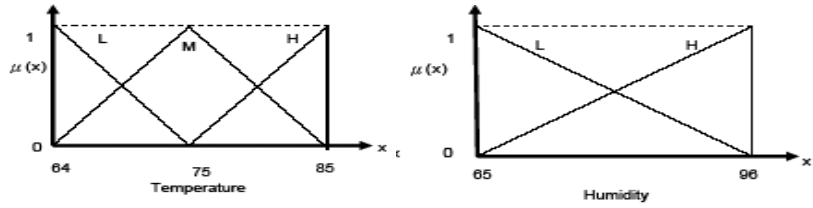

**Figure 6- Membership functions of Humidity and Tempreture[66].**

**Table 1-A snap shot of three objects of the data base [66].**

| data | outlook | temperature | humidity | windy | play | data |
|---|---|---|---|---|---|---|
| $X_1$ | sunny | 85 | 85 | false | no | $X_1$ |
| $X_2$ | sunny | 80 | 90 | true | no | $X_2$ |
|  |  |  |  |  |  |  |
| $X_{14}$ | rainy | 71 | 91 | true | no | $X_{14}$ |

The algorithm performs an iterative method to reduce the number of attributes. In each layer an attribute should be omitted. A notion of fuzzy entropy is applied for evaluating significant order of attributes which is determined as follows.

$$E(A) = \sum_{x \in X} \sum_{A_i \in A} \sum_{t_{ij} \in \mu_i} \frac{v_{ij}}{n} \times \mu_{ij}(x) \times \log_2 \frac{n}{v_{ij}}$$

where $\mu_{ij}(x)$ is the membership value of object x on the j-th linguistic term of i-th attribute( $t_{ij}$) and $v_{ij}$ is equals to the summation of $\mu_{ij}(x)$ for all x ∈ X, that is

$$v_{ij} = \sum_{x \in X} \mu_{ij}(x)$$

The fuzzy entropy is used to decide which attribute should be omitted for the construction of the upper layer concepts. The attribute which make the entropy smaller should be omitted. Table 2 shows 5 sets of attributes with 4 attributes are possible. Calculating the entropy for them shown in table 3, states that B1 should be selected.

**Table 2-5 number of 4 attribute sets**

| {B1} = {temperature, humidity, windy, play} | {B2} = {outlook, humidity, windy, play} |
|---|---|
| {B3} = {outlook, temperature, windy, play} | {B4} = {outlook, temperature, humidity, play} |
| {B5} = {outlook, temperature, humidity, windy} |  |

Table 3-The calculated entropy for eliminating one of the attributes in layer 5 [66].

| E({B5})=28.26 | E({B4}) =27.76 | E({B3}) =27.64 | E({B2}) =7.50 | E({B1}=27.23 |
|---|---|---|---|---|

The next part is merging the concepts of 5th layer for making the concepts of layer 4. The strategy for doing so, is merging concepts which have the same maximum linguistic value, for their linguistic variable. So the above example concept $C_2^5, C^5_{14}$ (objects corresponding to entity $x_2$ and $x_{14}$ in database) are merged as they both have Medium for their temperature, high for their humidity, T for windy and N for Play as their linguistic value with the maximum membership degree (Table 4).

For generating the membership of the combined concept, the minimum value is selected. So, concept $C_1^4$ has the following form.

$C1^4$ = (*, min(0/H, 0.5/H)+min(0.64/M, 0.5/M) +min(0.36/L,0/L), min(0.84/H, 0.81/H) +min(0.16/L, 0.19/L), min(1/T, 1/T)+ min(0/F, 0/F), min(0/Y,0/Y)+ min(1/N,1/N)).

Table 4-The linguistic values of database entities [66].

| play | Windy | Humidity | temperature | Outlook | data |
|---|---|---|---|---|---|
| 0/Y+1/N | 0/T+1/F | 0.65/H+0.35/L | 1/H+0/M+0/L | 0/O+0/R+1/S | $X_1$ |
| 0/Y+1/N | 1/T+0/F | 0.81/H+0.19/L | 0.5/H+0.5/M+0/L | 0/O+0/R+1/S | $X_2$ |
| | | | | | |
| 0/Y+1/N | 1/T+0/F | 0.84/H+0.16/L | 0/H+0.64/M+0.36/L | 0/O+1/R+0/S | $X_{14}$ |

Then $f(C_k^{l-1}, C_k^l)$ is computed to represent the degree of the sub-sumption of concept $C_k$ of Layer l-1 to the concept $C_k$ of layer l.

$$f(C_k^{l-1}, C_k^l) = \frac{\sum_{A_i \in B} \prod_{t_{ij} \in \mu_i}(1-|\mu_{ij}(C_k^{l-1})-\mu_{ij}(C_k^l)|)}{|B|}$$

Where $\mu_{ij}(C_l^k)$ is the membership value of the fuzzy concept $C_k^l$ to the j-th linguistic term of i-th attribute, $t_{ij}$.

There are other researchers on Fuzzy ontology generation from relational databases. Lv and colleagues [67] present a fuzzy ontology generation framework which generates fuzzy ontology from relational databases. The research does not talk about the method of mapping but about representing the fuzzy ontology. They propose fuzzy version of DLR called FDLR to represent fuzzy ontology. DLR is a special kind of description logic that has the ability to represent n-ary relations. Zhang and colleagues [68] propose an approach for constructing fuzzy ontologies from fuzzy UML models .Zhang and colleagues [69] propose a formal approach and an automated tool for constructing fuzzy ontologies from fuzzy Object-Oriented database (FOOD) models. Zhang and colleagues [70] propose an approach for constructing fuzzy ontologies from fuzzy relational data bases (FRDBs), and used the constructed fuzzy ontology to reason on FRDBs.

**6.3.2.2. Fuzzy Non-Taxonomic/Special Relations**

**6.3.2.2.1. FIM**

Lee and colleagues [5] define a 7-layer Fuzzy Inference Mechanism (FIM) for fuzzifying an existing crisp ontology. They use a layered ontology of the Chinese news domain, in which each concept has a membership degree for its belonging to different news events. In crisp ontology, each concept fully belongs to a special event or not, but in the fuzzy case each concept has membership value for its belongings to each event. The relation between events and concepts is a "part-of " (meronymy) relation.

The inputs of fuzzy inference mechanism are meaningful terms of the domain with their belongings to various events. The classification of the terms is done by a term classifier module. A preprocessing mechanism which uses Chinese news dictionary generates meaningful terms based on the news corpus.

Fuzzy inference mechanism follows a 7-layer architecture (Figure 7) to find the association between various terms of an event and the concepts. This association is then used to find membership degree of different concepts to different events.

Three fuzzy variables, including Term Part-of- Speech (POS), Term Word (TW) similarity, and Semantic Distance (SD) similarity- defined by domain experts-are used to find association between terms and concepts. POS compares differences in the part of speech of concepts and term. Term word similarity is a function of equal number of words in two terms. For computing the semantic distance similarity, the domain ontology is defined as five layers, including the *who* layer, *when* layer, *what* layer, *where* layer, and *how* layer. According to the layer which terms belong to, their semantic distance is calculated. All these variables are defined as fuzzy variables with three linguistic values low, medium, high. Figure 8 shows this fuzzy value for fuzzy variable POS.

A total criteria named term relation strength (TRS) is defined as a variable which integrates three other variables. Five linguistic values very-low, low, medium, high and very high are defined for TRS. Rules are defined for integrating values of POS, SD, TW to linguistic values of TRS. Table 5 shows some examples of the rules. For example, Rule 1 says that the output of POS-Low, TW-Low, SD-Low should be integrated to TRS-Very Low in Layer 4.

Figure 7 shows the 7-layer steps which FIM follows. First layer is "input linguistic layer" which accepts term set of events and concept set of d the omain as input and sends them to $2^{nd}$ layer. Next layer or "input term layer" performs the membership functions to compute the membership degrees of three variable POS, SD, TW. Third layer is "rule layer". There, each node contains a fuzzy rule, which shows the way of linking $3^{rd}$ layer nodes with the associated node of the fourth layer. "The output term layer" is the forth layer which performs the fuzzy OR operation to integrate the fired rules that have the same consequences to compute linguistic values of the fuzzy variable Terms Relation Strength (TRS). Fifth Layer, "The output linguistic layer" adopts the center of area (COA) method to perform the defuzzification process. This layer computes the TRS of the term and concept pair. The next two layers are summation layer and integration layer.

The summation layer performs the summation process over the terms of event for computing the TRS values of concept and event pair. The output of this layer contains belongings of each concept to event 1. Finally, the integration layer will integrate the membership degrees of the concept that belongs to all events of the domain ontology.

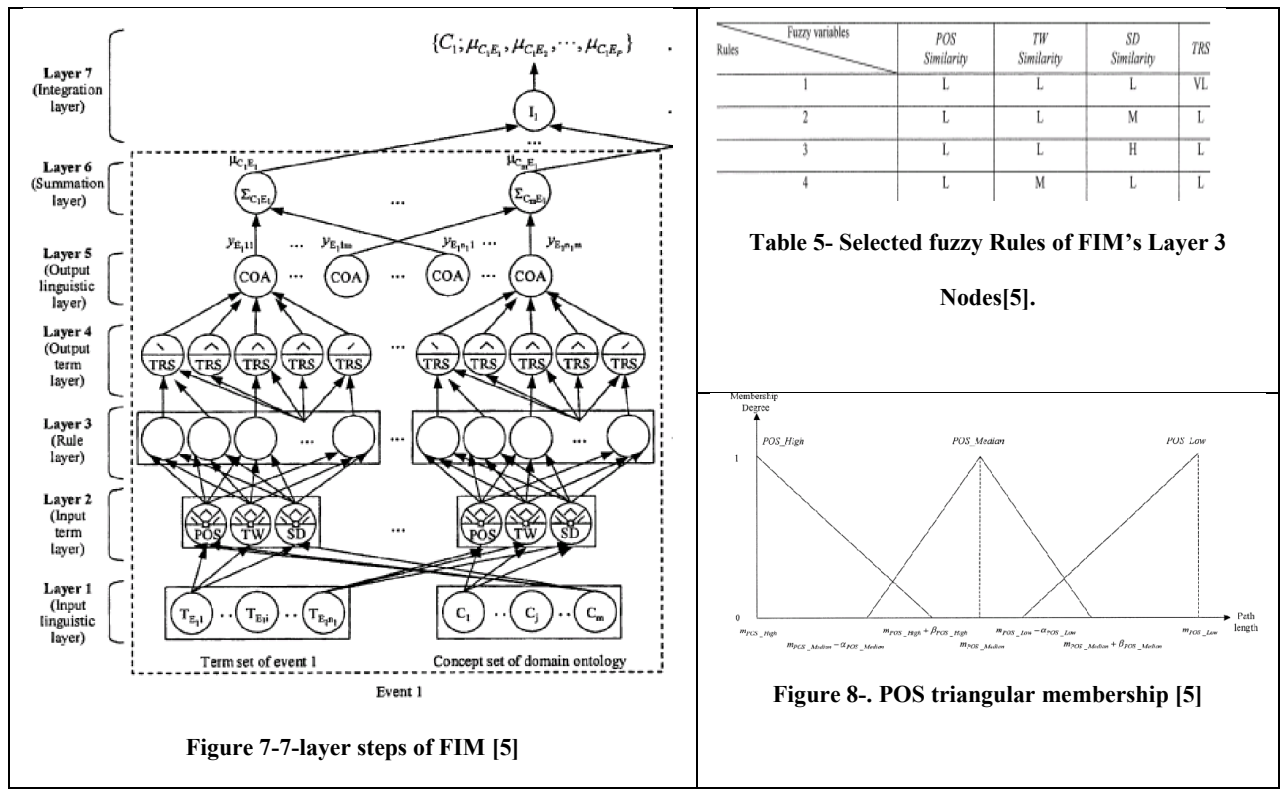

Figure 7- 7-layer steps of FIM [5]

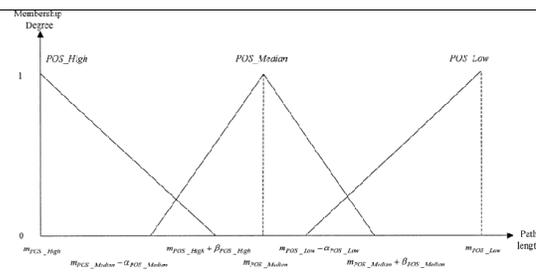

Table 5- Selected fuzzy Rules of FIM's Layer 3 Nodes[5].

Figure 8-. POS triangular membership [5]

### 6.3.2.2.2. Abulaish and Dey (B)

One of the researches on extracting non taxonomic fuzzy relations is the work of Abulaish and Dey [21]. They propose a mechanism for extracting fuzzy relations through text mining. Membership values of relations are functions of frequency of co-occurrence of concepts in the relations. This research works with GENIA ontology and tries to enhance it with fuzzy relations by using the information extracted from MEDLINE documents. The method consists of four modules.

The first two are doing a kind of preprocessing steps. The "Document Processor" module gets text documents which are tagged with ontology concepts and extracts sentences in them. "Relational Verb Extractor" is the $2^{nd}$ module which uses NLP techniques to mine all relational verbs from the document. "Biological Tag Association Extractor" is the fuzzyfying module. It computes association between pairs of tags. This module uses term frequency (TF) and inverse document frequency for computing weight $W_{ij}$ of tag $E_j$ in document $D_i$ as follows

$$W_{ij} = E_{ij} \times \log(N/n_j)$$

where $E_{ij}$ is the frequency of the jth entity $E_j$ in document $d_i$, $\log(N/n_j)$ is the inverse document frequency of entity $E_j$, N is the total number of documents in the collection $n_j$ is the number of documents that contain the $E_j$. Then the strength of association between a tag pair $E_j$ and $E_k$ is computed with the conjunction operator as

$$\mu(E_j, E_k) = \frac{\sum_{i=1}^{N} W_{ij} \otimes W_{ik}}{\sum_{i=1}^{N} W_{ij}}$$

where $\otimes$ denotes a fuzzy conjunction operator which is taken as a min operator in this case. A threshold value may be used to filter all associated fuzzy tag associations. The last module or "Fuzzy Biological Relation Extractor" identifies relations that have strength greater than specified threshold as "feasible fuzzy biological relations". So, the same relational verb may be associated to multiple entity-tag pairs with differing strengths

### 6.3.2.2.3. Parry

Parry [22] states another approach to fuzzify relations for solving the problem of overloaded concepts for query expansion in medical information retrieval. Overloaded concepts are concepts which occur more than once in the ontology in different locations (Figure 9-right). When a user searches about an overloaded concept the problem is to determine which location of the concept was intended by the user. The idea for clarifying this task is to subgroup users according to their interest and assigns fuzzy weight special for each group.

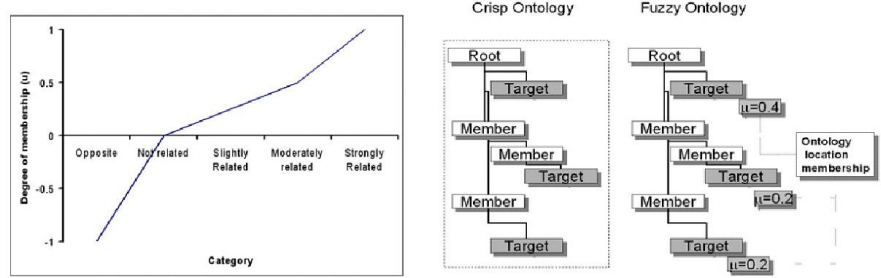

**Figure 9.** related term boxes (left), overloaded concept (right)[22]

In this approach, the ontology membership is normalized in a way that, for each of the terms in the ontology, sum of the membership values of the term be equal to 1. This is because of the methods' primarily concern which is mapping from queries to the ontology. So, for each term in a query, only one of the meanings is required which are exclusive. The fuzzification method has two steps. In the first step all locations of the overloaded concepts are assigned equal membership values in a way that sum of their memberships be equal to 1. In the second step these memberships are updated by analyzing the corpus of a certain domain or by using relevance feed-back from users that share particular common interests. The following equation is used to find new memberships.

$$\mu_{New} = \mu_{Old} \pm (\sqrt[2]{(\mu_i - \mu_{Old})^2 / Q_{Hist}})$$

The membership values of any other equivalent terms are decreased or increased in proportional amounts in order to maintain normality.

$$\mu_2(New) = \mu_2(Old) \mp \mu_{Change} \cdot \left(\frac{\mu_2(Old)}{\mu_2(Old) + \mu_3(Old)}\right)$$

Where $\mu_{Change}$ is the change in $\mu_1$. For example, consider 3 locations for a particular term, (L1, L2, L3) with membership values ($\mu_1=0.6$, $\mu_2=0.3$, $\mu_3=0.1$). If L1's membership value is decreased to 0.5 The new values are ($\mu_1=0.5$, $\mu_2=0.375$, $\mu_3=0.125$).

In analysis of the corpus, initial set of query terms is derived from the set of MeSH headings. These terms are fed into both GOOGLE (www.google.com) for the World Wide Web and PubMed (pubmed.gov) for Medline. The retrieved documents are processed. Each document retrieved was then searched for terms from the ontology. A weighting was introduced so that terms in the keyword section (PubMed) or meta tags (Google) were weighted as 3, terms in the title (PubMed) or headings (Google) were weighted as 2 and terms in the abstract (PubMed) or main body (Google) were weighted as one.

A parent or child of a test term are called it's "local term". The following equation is used to compute the membership function for each different location of the query terms.

$$\mu_{Automatic} = \frac{\sum_0^{i=k} L_i W_i}{\sum_0^{i=n} A_i W_i}$$

Where $L_i$ is the number of local terms discovered in each section of each document they are discovered, $W_i$ is the weighting, $A_i$ is the number of terms discovered and $W_i$ is the weighting.

In user feedback, the user performs a query and a set of documents is recovered. From the recovered documents the set of ontology terms are extracted and their affixes are removed. Then the user is asked to put them in the related boxes "Opposite", "Not Related", "Slightly Related", "Moderately Related" and "Strongly Related". Assigning the membership values of each term in each location is based on the membership function shown in Figure 9(left).

The query term is then compared to the terms in the "related terms boxes". A score is calculated for each potential meaning of each query term by summing the membership values of terms in the "related terms boxes" that are related to each potential location of the query term. To calculate the membership value of the query term in a particular location, the following equation is used:

$$\mu_{Result} = \frac{\sum_0^{i=n} \mu_i}{n}$$

Where $\mu_i$ is the membership value for each term the user has put into in the "related boxes". Only terms that are parents or children of the preferred meaning of the query term are included in this part of the calculation. If a term from the retrieved document occurs more than once, then each instance of the term is included. The value n is given by the number of such terms, including duplicates. If the calculation yields a membership value less than 0 then the value is reset to 0.

The result will be a fuzzy ontology that reflects the weights according to a user with a particular interest or document in a particular context of a domain. We call it context-dependent ontology. Similar approaches are applied in [71].

### 6.3.2.3. Fuzzy Taxonomic and Non-Taxonomic Relations

#### 6.3.2.3.1. FOGA
Tho and colleagues [25] propose Fuzzy Ontology Generation framework (FOGA) based on fuzzy set theory and formal concept analysis (FCA).

FCA defines formal contexts to represent relationships between objects and attributes in a domain. Formal concepts are generated from the formal contexts, this research proposes a combination of fuzzy logic and FCA as Fuzzy Formal Concept Analysis (FFCA), in which the uncertainty information is represented by a real number of membership value in the range of [0 1]. So, the definition of formal context and formal concept are changed to fuzzy formal concept and fuzzy formal context.

- *Fuzzy Formal Context:* A fuzzy formal context is a triple $K = (G, M, I(G \times M))$ where G is a set of objects, M is a set of attributes, and I is a fuzzy set on domain G×M. Each relation $(g, m) \in I$ has a membership value $\mu(g, m)$ in [0 1].

Each fuzzy formal context is a collection of objects called extent of the formal context and a collection of attributes as its intent. In contrast to crisp formal context, that each object has a special crisp value which is true(1) or false(0) for each attribute, in fuzzy formal context, each object then has a membership degree in [0 1] for each attribute. Table 6 shows an example of fuzzy formal context. Each document D1, D2, D3 has a membership value for each attributes "Data Mining"," Clustering ", "Fuzzy Logic".

**Table 6- Fuzzy Formal Context [25].**

|    | Data Mining | Clustering | Fuzzy Logic |
|----|-------------|------------|-------------|
| D1 | 0.8         | 0.12       | 0.61        |
| D2 | 0.9         | 0.85       | 0.13        |
| D3 | 0.1         | 0.14       | 0.87        |

- *Fuzzy Formal Concept:* Given a fuzzy formal context K = (G, M, I) and a confidence threshold T, The following two sets are defined

$A^* = \{m \in M | \forall g \in A: \mu(g, m) \geq T\}$ for $A \subseteq G$     $B^* = \{g \in G | \forall m \in B: \mu(g, m) \geq T\}$ for $B \subseteq M$

Fuzzy formal concept is a pair $A_f = (\varphi(A), B)$ where $B^* = A$, $A^* = B$

Each object $g \in \varphi(A)$ has a membership $\mu_g$ defined as: $\mu_g = \min_{m \in M} \mu(g, m)$ where $\mu(g, m)$ is the membership value between object g and attribute m defined in I. A formal concept is a collection of objects that have a collection of attributes in common and more than a predefined threshold T. Each object has a membership value to a special fuzzy formal concept that is the minimum of its membership to different attributes of that concept. Figure 10 shows the fuzzy formal concept of table 8 in contrast to the crisp formal concept.

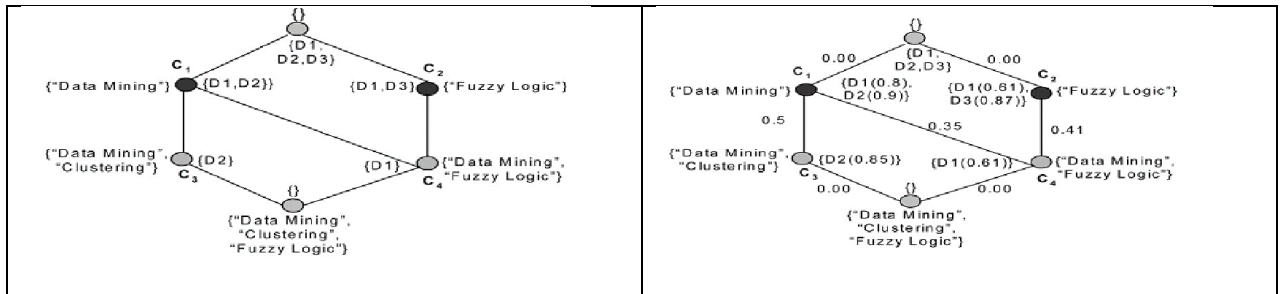

Figure 10-Formal concepts(left) and the fuzzy formal concept with confidence Threshold 0.5(right) [25]

By using the fuzzy definition of concepts, similarity of two concepts is defined as the similarity of their attribute set. The similarity of two fuzzy sets A, B is defined as $E(A, B) = \frac{|A \cap B|}{|A \cup B|}$

This similarity is then used for clustering of concepts. A conceptual cluster of a concept lattice K with a similarity confidence threshold T is a sub-lattice S(K) of K with two conditions.

1. It has a supremum concept Cs that is not similar to any of its super-concepts
2. Any concept C≠Cs in S (k) must have at least one super-concept C ∈ S (k) so that Similarity (C, C') > T

Each object has a membership value for each of the conceptual clusters that is the minimum of its' membership to concepts of that cluster.

Hierarchical relation is then defined between conceptual clusters. A conceptual cluster L1 is a sub-concept of a conceptual cluster L2, if its supremum is the sub concept of any concept $C' \in L_2$. Figure 11 Shows the conceptual clusters of Figure 10 with Threshold 0.4 and their hierarchical relations.

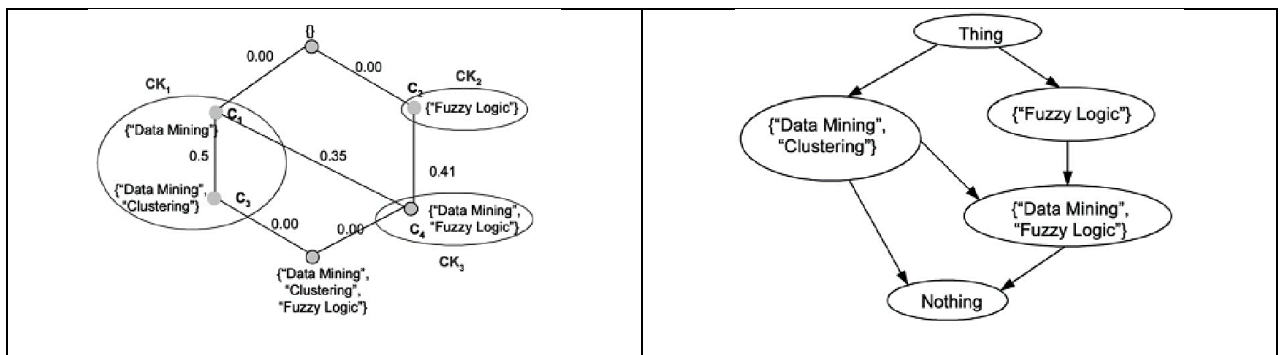

Figure 11-fuzzy conceptual clusters of the concepts generated in figure 10 with T=0.4. (left) ,Hierarchical relations between conceptual clusters(Right)[25].

This hierarchy is used to make fuzzy ontology as in the crisp formal concept analysis. Each Object has a fuzzy membership degree for all of its attributes, a fuzzy membership degree for being in a concept or conceptual cluster. Each conceptual cluster has a fuzzy membership degree for being a sub-concept of another conceptual cluster. So, in this framework all the parts are fuzzy.

FOGA also discusses about approximate reasoning for incremental enrichment of the ontology with new upcoming data. And a fuzzy-based technique for integrating other attributes of database to the ontology is proposed.

FOGA has been applied for constructing scholarly ontology from citation database [25], generating machine service ontology for semantic help-desk [72] and Reuters News Topic Themes Ontology. There are some other researches on fuzzy FCA [73] [74].

### 6.4. Developing Tools for Fuzzy Ontologies

There are some tools developed for fuzzy ontologies too. Some examples are discussed here.

Bobillo and Straccia [45] provide a Fuzzy OWL2 Protégé plug-in which uses OWL 2 annotation properties to encode fuzzy ontologies. Also they develop a parser for translating from OWL2 annotations representing fuzzy information into the language supported by some reasoners including fuzzyDL and DeLorean.

Calegari and Ciucci [26] extend the KAON Project to introduce fuzziness in ontology. A fuzzy inspector composed of a table representing fuzzy entity, membership degree and number of updates (Q) is developed. This developing tool is based on their method that lets updating fuzzy numbers by query.

Ghorbel and colleagues [75] introduce a fuzzy plug-in for Protégé 3.3.1.The plug-in allows the definition of parameterized membership functions and gives support to instantiate fuzzy concepts and roles. Also, it allows automatic computing of membership degrees and querying fuzzy ontologies based on fuzzy criteria.

Slavíček [76] provides a library to integrate a fuzzy ontology with object-oriented programming (OOP) classes written in .NET. The implementation currently supports FuzzyOWL2 ontologies with FuzzyDL reasoned, but it can be modified to support any fuzzy ontology notation and fuzzy reasoner.

### 7. Where Fuzzy Ontology Is Used?

All the applications that handle vague knowledge may use fuzzy ontologies. One of the most important applications of fuzzy ontology is semantic web [25],[72] ,[77]. For example Tho and colleges [72] use a fuzzy machine service ontology in a semantic help-desk for supporting customer services over the semantic web environment. Zhai and colleagues [78] use fuzzy ontology to exchange and integrate fuzzy systems knowledge with other Semantic Web applications. It has been applied in social network content analysis which has been

Another important application is information retrieval. [65], [79],[80],[81],[82]. As an example Parry [71] applies a fuzzy ontology in medical information retrieval. Zhai and colleagues use a fuzzy ontology for semantic query expansion in electronic commerce [34].Widyantoro and Yen [62] use an ontology of term associations for query refinement. Calegari and Loregian [79] use a special type of non-taxonomic fuzzy relations, called correlations for information retrieval. Lau and colleagues [65] use fuzzy ontology for the estimation of semantic granularity of documents to improve the effectiveness of IR.

Some of the other applications which use fuzzy ontologies include news summarization [5], meaning-based NLP interpretation [83], data mining [84], personal diabetic diet recommendation [85], visual video content analysis and indexing [86],[4], document clustering [87], knowledge extraction [88], group decision making systems [4] and fuzzy system modeling[89].

Fuzzy ontologies also have been applied in different domains to represent uncertain knowledge, such as the domain of educational computer games[58], the knowledge in an intelligent multi-cascade control system [90], fuzzy ontology of food [91] personal profile ontology [92], personal multimedia information [93], fuzzy ontology of computer threats [94].

Fuzzy ontology has application in the area of ontologies too. Qui et al. [95] define a semantic similarity assessment based on fuzzy weights for modularization of ontologies. Abulaish and Dey [30] define a special metric called consistency which shows how consistent each concept is defined among different ontologies. Consistency is computed using a function of fuzzy weights of the relations of an entity to other concepts. It is used for providing interoperability among ontologies.

## 8. How Fuzzy Ontology Is Evaluated?

For evaluating fuzzy ontologies, most of the researchers evaluate the application which contains fuzzy ontologies or use the usual methods of evaluating crisp ontologies. To the best of our knowledge, the only research which is explicitly about evaluating fuzzy ontologies is the work of Ivanova [96].He propose a special metric defined in this way .

$$K=\sum_{(Pi\neq 0)} pi/n - \sum_{pi=0} 0.1/n - \log(n_1/n)$$

Where $p_i$ is the value of *probability of* (sub) property of ontology element, which means what is the probability that this element locates in a right position, n is the number of all ontology terms, $n_1$ is the number of all valuable domain terms, extracted from selected text documents, that aren't ontology terms. Increasing metric values stands for well working of the learning process while decreasing means that it does not work properly. He does not talk about how the probability and user agreements are estimated.

Also in [60] some criteria including correctness, accuracy, completeness and consistency are introduced for evaluating fuzzy ontology. The metric are discussed in section 6.1.1.

## 9. Comparison

All fuzzy ontology developing methods fuzzify some ontology elements according to their application needs. The fuzzy membership degrees are obtained from a source of knowledge which may be a classified term set or ontology based tagged document, etc. Having the source of knowledge, a preprocessing step may be needed to prepare the knowledge for computing the fuzzy membership degrees. Finally, the fuzzification is done with different methods. Thus, according to what they have and what they need, methods use different approaches for their fuzzification, and they are different in what they do, and how they do that.

For comparing these methods, we adapt the framework introduced by Shamsfard and Abdollahzadeh [12] with some modifications. We change the dimension "learning element" to "fuzzy element" and we add fuzzy reasoning dimension to the framework. The dimension "Test and evaluation Environment" is also changed to "application-domain". Thus, we introduce a framework for comparing fuzzifng approaches according to the following six dimensions.

- *Fuzzy Element:* Learning methods are different in their fuzzy element. Some of them make fuzzy taxonomic relations such as method in PASS, some make fuzzy non-taxonomic relations such as FIM. Some have fuzzy property descriptors as method proposed in Abulaish and Dey (A).

- *Starting point:* This dimension talks about where the method begins the developing (what it has as the starting knowledge). It includes two parts. Input and starting knowledge. Developing methods are different in their starting knowledge; some methods have a crisp ontology for starting such as method in the Abulaish and Dey(A) and some others don't have it, such as method in Lau and colleagues [24] . Some of them use a special kind of lexicons like WordNet such as PASS and others such as FIM may have a dictionary of words.  Social network data have recently been applied as a source of input for fuzzy ontology generation as well [97]; which has a wide range of application in other areas of Artificial Intelligence (e.g. [98]). Furthermore, methods are different in their input. They may use text or Input type may be structured, semi structured or unstructured. The input language is also different.

- *Pre-processing:* This dimension answers if a pre-processing step is needed for transforming the input to the format for the beginning of learning. For example FOGA works with a cross table of words in the domain so a preprocessing for preparing the input is required or Lau and colleagues [24] does a linguistic preprocessing (pos tagging, stemming, stop word removal).

- *Result:* This dimension discusses about how rich the result of a method is. Some of them give methodology for developing fuzzy ontology manually, some other methods give a model or framework as their result (we mention them as manual or semi-automatic methods in this paper) and others give ontology as their result, which are different in their type, structure and representing language. The type of the ontology may be domain specific, general or context dependent. Some has layered structure. The representation language for the developed fuzzy ontology is also different.

- *Learning method:* Their learning methods are also different in their process, approach and level of automation. The approach may be statistical, linguistic, rule based or fuzzy. The process will also be different. Some do clustering, others may use statistical analysis. Automation levels are fully automatic, semi-automatic or manual. The special point about learning methods is that when we are talking about learning methods we mean just learning fuzzy memberships. All the other operations are gone to the pre-processing step. For example, Abulaish and Dey(B) make fuzzy non-taxonomic relations with respect to a crisp ontology. Some linguistic operations are done for tagging the documents. These steps are taken as a part of pre-processing dimension.

- *Application and the domain/ Evaluation and test environment:* This dimension talks about the application and the domain the fuzzy ontology is applied to.

- *Fuzzy reasoning*: Most of these methods are using fuzzy weights to make fuzzy ontologies and don't provide fuzzy reasoning on it. But a few support some kinds of reasoning such as FOGA which provides approximate reasoning and Gu and colleagues [61] which talks about a description logic which fits their model.

- In Table 7[3].we compare learning methods discussed in this paper according to these dimensions.

---

[3]none means it does not have a meaning here, not available means it was not discussed, not specific means it makes no difference.

**Table 7. Comparison of developing methods**

| | Fuzzy Element | Starting point | | | Pre-Processing | Result | | | Learning method | | | Evaluation and Test/Domain and application | Fuzzy reasoning |
|---|---|---|---|---|---|---|---|---|---|---|---|---|---|
| | | Starting-Knowledge | | | | Type | Structure | Representing languages | Automation level | Process | Approach | | |
| | | Input | Type | | | | | | | | | | |
| | | | | Language | | | | | | | | | |
| FOGA | Instances, Relations, Attributes | None | | | Make a cross table of words | General | Not-Specific | Owl | Automatic | Clustering FCA | fuzzy | Machine Service Ontology, scholarly ontology, Reuters News Topic Ontology | √ |
| | | cross-table of words | English | | | | | | | | | | |
| | | | Semi-structured | | | | | | | | | | |
| FIM | meronymy relations | Chinese news Dictionary + Domain Ontology | | | Classifying of terms by term class | Domain Specific | Layered | Owl | Automatic | 7 layer fuzzy, defuzzy | Combinatorial (Linguistic, rule base, fuzzy) | news Summarization | - |
| | | classified meaningful terms | Structured | | | | | | | | | | |
| | | | Chinese | | | | | | | | | | |
| Abulaish And Dey(A) | Property descriptors | Domain Ontology | | | Pattern filtering | Weighted Domain Ontology | Not-Specific | owl | Automatic | None | Combinatorial (Linguistic, rule based) | IR, Inter-operability | |
| | | Text Documents | unstructured | | | | | | | | | | |
| | | | English | | | | | | | | | | |

| Name | Relations | Domain Ontology | | Preprocessing | Output | Representation | Format | Learning | Reasoning | Evaluation | Application | Inference |
|---|---|---|---|---|---|---|---|---|---|---|---|---|
| Abulaish and Dey(B) | Non-taxonomic relations | ontology based tagged Medline documents | Semi-Structured / Scientific English | Shallow NLP pre-processing + ontology tagging | Weighted domain Ontology | Not-Specific | Not-Available | Automatic | Using Conjunction operator | Combinatorial (linguistic +statistical) | Information extraction, retrieval on biologic domain | - |
| 3-Layer | Property-values | linguistic variable ontology | Structured / Not-Specific | None | Model | Layard | RDF/OWL | Semi-automatic | None | None | Semantic query IR in Ecommerce, SCM, Transport, | - |
| Gu | Relations | Candidate FCBR classes | Structured / Not-Specific | None | Framework | Not-Specific | EF-SHIN fuzzy DL | Semi-automatic | None | None | Not-Available | √x |
| FOM | Relations | Structured relation degree between classes / Not-Specific | | None | Framework | Graph(Matrix) | Xml +OWL | Semi-Automatic | Inferring class memberships | Logical | Personal profile information | formal reasoning |
| Parry | Relations of overloaded concepts | Domain Ontology | | Not-Available | Context Dependent Ontology | Weighted Domain Ontology | Not-Available | Semi-Automatic | Updating By Query(analyzing the corpus or user feedback) | Statistical, | Medical IR | - |
| | | Textual corpus | Unstructured | | | | | | | | | |
| | | | English | | | | | | | | | |
| Yanhui | Relation | None | | None | Ontology | Not-Available | FDLR | automatic | Not-Available | Not-Available | Not-Available | - |
| | | Data Base Object | Structured | | | | | | | | | |
| | | | Not-Specific | | | | | | | | | |
| Lau and Hsu | Taxonomic relation | WorldNet | | Linguistic, pattern filtering | Domain specific-Ontology | Hierarchy of context vectors | Owl/RDF | automatic | Conjunction operator +2 steps of pruning | statistical | Query expansion IR, concept map generation | - |
| | | Textual Data base | Unstructured | | | | | | | | | |
| | | | English | | | | | | | | | |
| PASS | Taxonomic Relation | WordNet | | Linguistic, pattern filtering | Domain specific Ontology | Hierarchy of terms | Not-Available | automatic | Conjunction operator +2 steps of pruning | statistical | Query refinement | - |
| | | Association Terms | Unstructured | | | | | | | | | |
| | | | English | | | | | | | | | |

| | | | | | | | | | | | | | |
|---|---|---|---|---|---|---|---|---|---|---|---|---|---|
| Chein | Taxonomic Relation | None | | None | Not-specific | Hierarchy of Concepts | Not-Available | automatic | clustering | Fuzzy | Not-Available | - |
| | | Data Base Objects | Structured | | | | | | | | | |
| | | | Not-Specific | | | | | | | | | |
| Samani and ShamsFard | Properties | Domain Ontology | | General | Model | Not-specific | E-owl | Semi-Automatic | None | None | Qualitative spatial reasoner | √ |
| | | Linguistic values, modifiers, membership functions | Semi-structured | | | | | | | | | |
| | | | Not specific | | | | | | | | | |
| IKARUS-Onto | Properties and relations | Domain Ontology | | General | methodology | not-specific | not-specific | manual | None | None | Enterprise ontology for a consulting firm | - |
| | | Fuzzy membership degrees | structured | | | | | | | | | |
| | | | None | | | | | | | | | |

## 10. Conclusion and Future Work

In this paper, we surveyed the field of fuzzy ontology and its development. It was stated that fuzzy ontologies are needed because of the vagueness inherent in some real-world domains, the requirement of some applications and differences in expert's conceptualization of a domain. Then some different definitions of fuzzy ontologies were proposed. It was discussed that due to the requirements of the application which uses fuzzy ontologies, different elements could get fuzzy. A comprehensive definition for the fuzzy ontology was provided too.

Then, fuzzy description logic was discussed as a theoretical counterpart of fuzzy ontologies and introduced as a good candidate for representing and reasoning of fuzzy ontologies. Anyway, none of the representation methods are a standard one.

After that, different approaches for developing fuzzy ontologies were discussed and compared according to the proposed framework. The proposed framework compares developing methods based on their fuzzy element, starting point, preprocessing, result, learning method, application and evaluation domain and supporting of fuzzy reasoning.

However, there are still no fully automatic methods for developing fuzzy ontologies. Most of the developing methods are domain specific. They make little sense in directing the construction of fuzzy ontologies in other domains. Most of them use statistical method for fuzzification and little rule based or linguistic methods are used for fuzzification. They fuzzify elements according to their application requirements. Some methods use a crisp ontology as a starting knowledge and the fuzzification method is dependent to the structure of the underlying crisp ontology. Most of them do not contain fuzzy entailment or fuzzy representation of ontologies.

One of the most important applications of fuzzy ontologies is qualitative (approximate) reasoning. For a system to be able to reason with quality, having a fuzzy ontology is a good idea. However, there should be a method which can map quantitative element to their qualitative parts.

Finally, some applications of fuzzy ontologies and researches on evaluating them were introduced.